\documentclass{article}
\usepackage{spconf,amsmath,graphicx,hyperref}
\usepackage{booktabs}
\usepackage{multirow}
\usepackage{cite}
\usepackage{etoolbox}
\makeatletter
\patchcmd{\thebibliography}
  {\usecounter{enumi}}
  {\setlength{\itemsep}{-0.4em}\usecounter{enumi}}  
  {}{}
\makeatother
\usepackage{enumitem}
\usepackage{algorithm}
\usepackage{algpseudocode}
\usepackage{arydshln}

\usepackage{url}
\title{KG2QA: Knowledge Graph-Enhanced Retrieval-Augmented Generation for Communication Standards Question Answering}
\name{Zhongze Luo$^1$, Weixuan Wan$^2$, Tianya Zhang$^1$, Dan Wang$^2$, Xiaoying Tang$^{1*}$\thanks{$^*$Corresponding author: Xiaoying Tang.}}
\address{$^1$School of Science and Engineering, The Chinese University of Hong Kong, Shenzhen, China\\
$^2$School of Microelectronics, Xi'an Jiaotong University, China
}
\begin{document}
%
\maketitle
\begin{abstract}
The rapid evolution of communication technologies has led to an explosion of standards, rendering traditional expert-dependent consultation methods inefficient and slow. To address this challenge, we propose \textbf{KG2QA}, a question answering (QA) framework for communication standards that integrates fine-tuned large language models (LLMs) with a domain-specific knowledge graph (KG) via a retrieval-augmented generation (RAG) pipeline. We construct a high-quality dataset of 6,587 QA pairs from ITU-T recommendations and fine-tune Qwen2.5-7B-Instruct, achieving significant performance gains: BLEU-4 increases from 18.86 to 66.90, outperforming both the base model and Llama-3-8B-Instruct. A structured KG containing 13,906 entities and 13,524 relations is built using LLM-assisted triple extraction based on a custom ontology. In our KG-RAG pipeline, the fine-tuned LLMs first retrieves relevant knowledge from KG, enabling more accurate and factually grounded responses. Evaluated by DeepSeek-V3 as a judge, the KG-enhanced system improves performance across five dimensions, with an average score increase of 2.26\%, demonstrating superior factual accuracy and relevance. Integrated with Web platform and API, KG2QA delivers an efficient and interactive user experience. Our code and data have been open-sourced \footnote{\url{https://github.com/luozhongze/KG2QA}}.

\end{abstract}
\section{Introduction}
\label{sec:intro}
The rapid advancement of technology in the communication sector has led to an exponential increase in communication standards \cite{kaushik2024toward}. Traditionally, consulting these standards requires significant expertise and manual intervention, often involving extensive periods for information retrieval and interpretation. Such conventional methods are inefficient and unable to keep pace with rapid technological developments, necessitating smarter, more automated approaches for consulting and knowledge management \cite{10890582}.

Recent developments in artificial intelligence, particularly large language models (LLMs), have demonstrated remarkable capabilities in understanding and generating human-like text across general domains \cite{ge2023openagi}. However, their performance in specialized fields, like communication standards, remains limited due to the sparse representation of specialized terminologies and domain-specific knowledge in pretraining datasets \cite{demszky2023using,han2025synthetic}. To address this limitation, fine-tuning methodologies such as Low-Rank Adaptation (LoRA) have been employed, significantly enhancing the semantic comprehension and contextual accuracy of LLMs in niche areas \cite{lin2024data,sun2024lawluo,wu2025llm}.

Complementing LLMs, knowledge graph (KG) offers a structured approach to represent complex, interconnected domain-specific knowledge \cite{chen2020review,sun2024forpkg,10889242}. Particularly in narrowly defined domains, such as communication standards, domain-specific KG facilitates efficient and precise information retrieval \cite{yu2021kg,yu2021knowledge,zirui2021survey}. By constructing a carefully designed ontology structure and leveraging LLMs for entity and relationship extraction, high-accuracy KG can be established, markedly improving retrieval performance \cite{bogdanov2024nuner,kommineni2024human,sun2025compliance,fan2024survey,zhang2024raft,10445901,10446669}.

\begin{figure}[!t]
    \centering
    \includegraphics[width=0.9\linewidth]{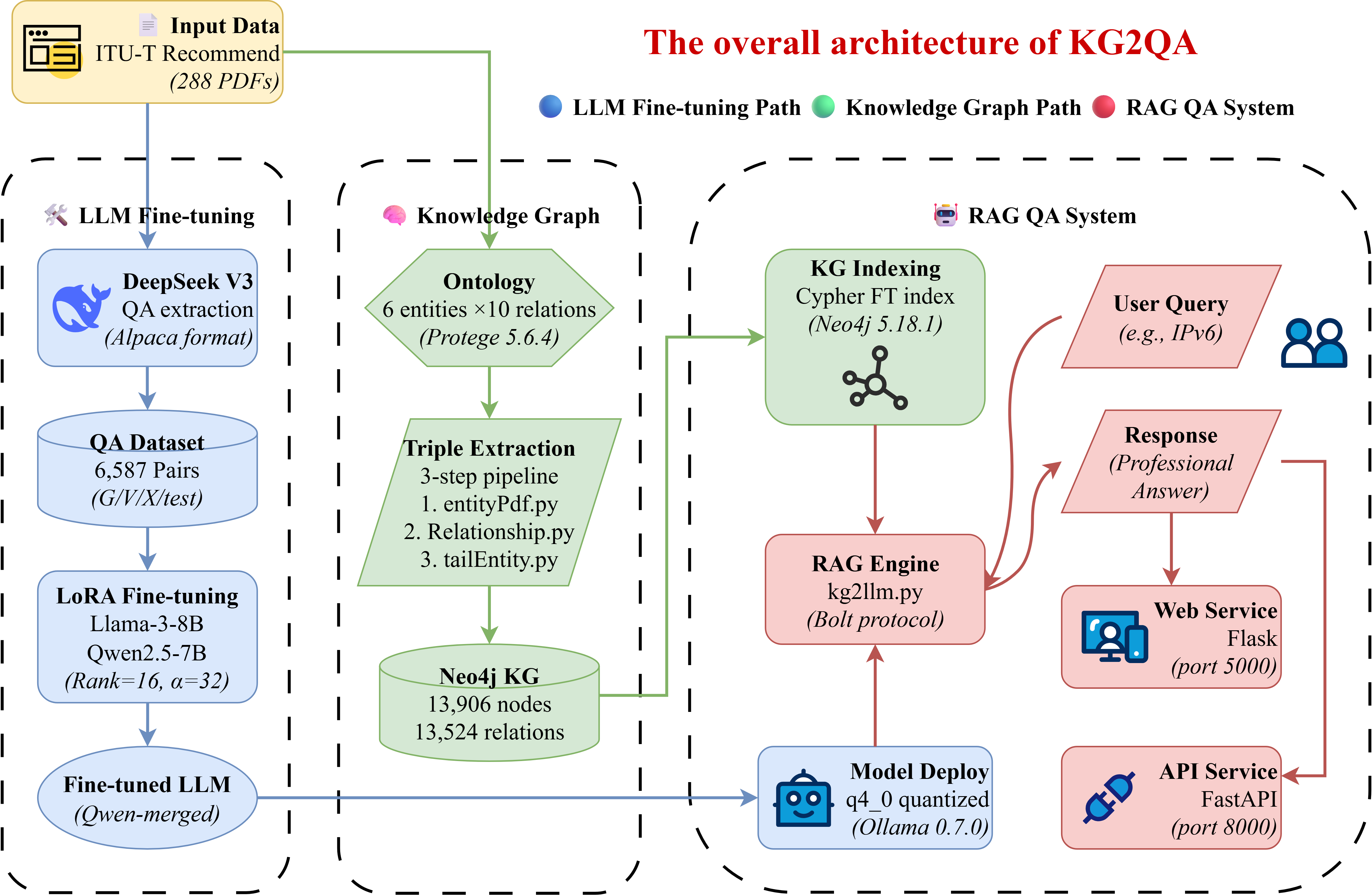}
    \caption{The overall architecture of KG2QA.}
    \label{fig:architecture}
\end{figure}

We propose an integration \textbf{KG2QA} framework, which combines domain-specific KG and fine-tuned LLMs using retrieval-augmented generation (RAG) methodologies, and we apply it to the field of communication standards question answering (QA). Our approach enhances intelligent consulting services by achieving precise and swift query responses, enriched by structured domain-specific knowledge. Experimental results demonstrate substantial improvements in query accuracy, response speed, and overall user interaction experience compared to traditional and general-purpose AI consulting methods. The overall architecture of KG2QA is shown in Fig. \ref{fig:architecture}. In summary, the main contributions are as follows:

\begin{itemize}[nosep, leftmargin=*]
\item We perform efficient domain adaptation of LLMs using LoRA fine-tuning for communication standards QA tasks.
\item We construct a structured KG for communication standards based on LLM-assisted triple extraction.
\item We develop a RAG-based QA system integrating the fine-tuned LLMs with the domain-specific KG.
\item We provide fully open-sourced, reproducible KG2QA framework and dataset for domain-specific QA in communication standards, and evaluate the improvements.
\end{itemize}

\section{Methods}
\label{sec:Methods}

\subsection{Fine-tuning}

The dataset used for fine-tuning LLMs is in the field of communication standards, and the language is English. The data is derived from ITU-T Recommendations and is taken from three series of recommendations. A total of 288 PDF source files are selected from the three series for QA construction. The dataset is divided into the training set and the test set. The training set consists of data of three categories (G, V, X), and the test set consists of data of one category (Test), both of which are saved in JSON format. The DeepSeek-V3 API is used to perform QA extraction on the source PDF files to obtain the dataset. The format is all in the standard Alpaca format, which is an instruction-supervised fine-tuning dataset. The data volume is shown in Table \ref{tab:data_volume}.

\begin{table}[!htbp]
    \centering
    \caption{Data volume description.}\label{tab:data_volume}
    \begin{tabular}{cccccc}
        \toprule
        Category & G & V & X & Test & All\\
        \midrule
        QA Quantity & 898 & 1,101 & 3,779 & 809 & 6,587 \\
        \bottomrule
    \end{tabular}
\end{table}

\subsection{KG Construction}

\begin{table}[!htbp]
    \caption{Entity type definitions.}
    \centering
    \resizebox{\linewidth}{!}{
    \begin{tabular}{ccc}
        \toprule
        Entity type & Identifier & Description \\
        \midrule
        Identifier & IDEN & Establish a unique identifier for an entity, \\
        & & including its common name, type, and \\
        & & criteria for defining its source \\
        Structure/Composition & STR\_COM & Describe how the internal units, \\
        & & components or the whole of an entity are \\
        & & constituted \\
        Suitability/Context & APP\_CON & Define the context, scope, constraints or \\
        & & specific context of the application entity \\
        Action & ACT & The operational steps or behaviors carried \\
        & & out or involved \\
        Value & VALUE & Specific values, encodings or Settings \\
        & & associated with an entity \\
        Function & FUN & The specific action or function performed \\
        \bottomrule
    \end{tabular}
    }
\label{tab:entity_type_def}
\end{table}

\begin{table}[!htbp]
    \caption{Relationship type definitions.}
    \centering
    \resizebox{\linewidth}{!}{
    \begin{tabular}{ccccc}
        \toprule
        Relationship & Identifier & Domain of & Range of & Inverse \\
        type & & definition & value & relationship \\
        \midrule
        Contain & contain & STR\_COM/ & IDEN/ & Be contained \\
        & & IDEN & VALUE & (isContained) \\
        Rely on & isReliedOn & FUN & VALUE & / \\
        Accomplish & accomplish & ACT & FUN & / \\
        Limit & limit & STR\_COM/ & FUN/ACT & / \\
        & & APP\_CON & & \\
        Relevant & relevant & FUN/ & IDEN/ & This relationship \\
        & & VALUE/ & APP\_CON & is self-inverse \\
        & & ACT/IDEN & & \\
        Execute & execute & VALUE & ACT & Undo (undo) \\
        Influence & influence & APP\_CON & STR\_COM & Be influenced \\
        & & & & (isInfluenced) \\
        \bottomrule
    \end{tabular}
    }
\label{tab:relationship_type_def}
\end{table}

In the process of ontology design, the domain characteristics of communication standards are fully considered. We adopt the KG construction process based on LLMs, which is divided into three steps in total: head entity recognition, latent relationship recognition, and tail entity recognition. Finally, a complete triple is extracted and the KG is uploaded to the Neo4j platform for visualization and evaluation. We design 6 types of entities: Identifier, Structure/Composition, Suitability/Context, Action, Value and Function, as shown in Table \ref{tab:entity_type_def}. After defining the entity types, we define 10 types of potential relationships, as shown in Table \ref{tab:relationship_type_def}.

\begin{algorithm}[!t]
\small
\caption{End-to-End KG Construction}\label{algo:kg-construction}
\textbf{Input:} $\texttt{PDF\_DIR}$: Directory containing the original PDF files,\\ $\theta$: Confidence threshold \\
\textbf{Output:} $KG[f]$: Complete KG triples

\begin{algorithmic}[1]
\For{each file $f \in \texttt{PDF\_DIR}$}
    \State $text_f \gets \texttt{extract\_text\_from\_pdf}(f)$
    \State $H[f] \gets \texttt{LLM}(\texttt{head\_entity\_prompt}(text_f))$
    \For{each $h \in H[f]$}
        \State $ctx \gets h.context$
        \State Append $(h, \texttt{LLM}(\texttt{relation\_prompt}(h, ctx)))$ to $R\_T[f]$
    \EndFor
    \For{each $(h, rels) \in R\_T[f]$}
        \For{each $(r, t') \in rels$}
            \If{$r.confidence \geq \theta$}
                \State $t\_type \gets \texttt{determine\_tail\_entity\_type}(t'.name, ctx)$
                \State Append $(h, r, t', t\_type)$ to $KG[f]$
            \EndIf
        \EndFor
    \EndFor
    \State \texttt{save\_to\_json}($KG[f]$, \texttt{"final\_triples\_f.json"})
\EndFor
\end{algorithmic}
\end{algorithm}

We design a pipeline for extracting KG triples from PDF files based on LLMs, which has a clear division of labor and modularly realizes the extraction, as shown in Algorithm \ref{algo:kg-construction}.

The process begins with a single PDF file, denoted as $f$. From this file, we extract its entire textual content, represented as $text_f$. Based on this text, the algorithm identifies a list of head entities, which is stored in $H[f]$. For each head entity $h \in H[f]$, we define its surrounding text as its context, $ctx$. Next, the algorithm leverages the head entity $h$ and its context $ctx$ to extract potential relationships $r$, and candidate tail entities $t'$. For a given file $f$, the collection of all extracted candidate relations and tail entities is stored in $R\_T[f]$. To ensure the quality of the extracted facts, we employ a confidence threshold, $\theta$ (e.g., 0.8), to filter out relations with low confidence scores. For each tail entity $t'$ that passes this filter, we further determine its attribute type, denoted as $t\_type$. Finally, all validated KG triples, consisting of a head entity $h$, a relation $r$, and a tail entity $t'$, are aggregated into $KG[f]$. This structure, $KG[f]$, represents the final KG constructed from the single source file $f$.

\subsection{QA System}

\textbf{Ollama model remote access to KG.} The fine-tuned Qwen Merged model is converted to GGUF format, quantized (q4\_0), and deployed locally using Ollama (v0.7.0). We enable remote access to the local Neo4j instance from the server running the Ollama model via the Bolt protocol (port 7687). The QA system first retrieves relevant context from the KG based on user keywords, then feeds this context to the fine-tuned Qwen Merged model to generate answers.

\begin{figure}[!htbp]
    \centering
    \includegraphics[width=0.9\linewidth]{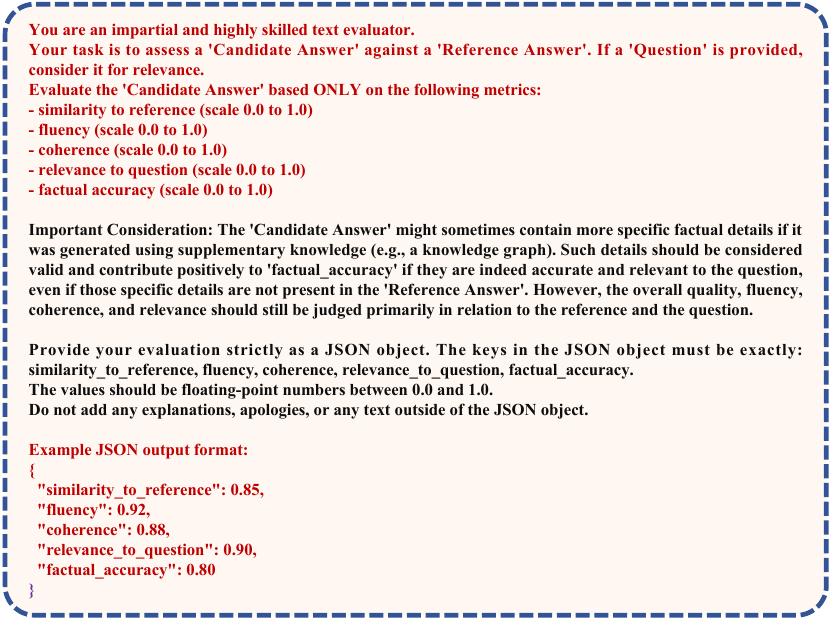}
    \caption{Prompt of LLM-based judge.}
    \label{fig:prompt}
\end{figure}

\textbf{Quantitative evaluation of answer quality using a LLM-based judge.} To quantitatively assess the performance improvement from the KG, we adopted the ``LLM as a Judge''. This approach overcomes the limitations of lexical-overlap metrics like BLEU/ROUGE, which may unfairly penalizes a RAG-generated answer for containing more correct facts than the static reference. We use the DeepSeek-V3 API as an impartial judge to score answers from a baseline LLM-Only system and our LLM+KG system on five dimensions. Prompt is shown in Fig. \ref{fig:prompt}.

\section{Results and Analysis}

\subsection{Fine-tuning Experiments}

This paper's server Linux version: Ubuntu 22.04, using NVIDIA® GeForce RTX™ 4090D 24G, Fine-tuning was carried out using the LLaMA-Factory framework, and a control experiment was conducted with the same parameter settings for the two base models. The models are Llama-3-8B-Instruct and Qwen2.5-7B-Instruct respectively. Llama-Factory version: 0.93.dev0, Python version: 3.10.0, PyTorch version: 2.6.0+cu124, CUDA version: 12.8. The LoRA fine-tuning parameters are set to lora\_rank: 16, lora\_alpha: 32, learning\_rate: 5.0e-5. The evaluation results of the base model and the merged model on the test set are shown in Table \ref{tab:eval_base_combined}.

\begin{table}[!htbp]
\centering
\caption{Evaluation results in fine-tuning.}
\resizebox{\linewidth}{!}{
    \begin{tabular}{ccccc}
    \toprule
    Metrics & Llama Base & Llama Merged & Qwen Base & Qwen Merged \\
    \midrule
    BLEU-4 & 37.3405 & 66.4780 & 18.8564 & \textbf{66.8993} \\
    ROUGE-1 & 48.2548 & 70.2607 & 35.8729 & \textbf{70.9748} \\
    ROUGE-2 & 25.2638 & 51.7437 & 16.0076 & \textbf{52.9495} \\
    ROUGE-L & 31.6350 & \textbf{70.7144} & 18.7772 & 61.4781 \\
    predict runtime & 0.0037 & 0.0038 & \textbf{0.0036} & 0.0037 \\
    samples per second & 0.2790 & 0.7720 & 0.1530 & \textbf{0.8700} \\
    steps per second & 0.1400 & 0.3870 & 0.0760 & \textbf{0.4360} \\
    runtime & 2900.5805 & 1047.5824 & 5301.4163 & \textbf{929.6552} \\
    \bottomrule
    \end{tabular}
}
\label{tab:eval_base_combined}
\end{table}

The model after fine-tuning and merging is superior to the base model in most indicators. Taking the Qwen series as an example, the BLEU-4 score of the combined model increased from 18.8564 to 66.8993, and ROUGE scores also improved significantly. In terms of efficiency indicators, the reasoning time of the combined model has been significantly shortened. The time required for one inference of the Qwen base model has decreased from 5301.4163 seconds to 929.6552 seconds, samples per second and steps per second have increased to 0.8700 seconds and 0.4360 seconds respectively, the Llama series also shows a similar trend. Therefore, the fine-tuned merged model is superior to the base model in both generation quality and reasoning efficiency. The Llama base model outperforms the Qwen base model, but after fine-tuning, the Qwen Merged model is overall superior to the Llama Merged model, which indicates that although the Llama base model has better native performance, the Qwen model is more adaptable to the data of this fine-tuning task.

We also select the mainstream models such as DeepSeek, Kimi, Doubao, ChatGPT, and Gemini on the market for the metric evaluation on the test set, comparing them with the fine-tuned Llama Merged and Qwen Merged model. The evaluation results are shown in Table \ref{tab:eval_qwen_general}.

\begin{table}[!htbp]
    \centering
    \caption{Evaluation results with other models.}
    \resizebox{\linewidth}{!}{
    \begin{tabular}{lcccc}
        \toprule
        Models & BLEU-4 & ROUGE-1 & ROUGE-2 & ROUGE-L \\
        \midrule
        DeepSeek   & 37.4556 & 47.0695 & 27.2858 & 36.3754 \\
        Kimi       & 59.9932 & 39.4126 & 25.0124 & 31.9145 \\
        Doubao     & 28.9278 & 55.0455 & 37.9567 & 46.3878 \\
        ChatGPT    & 35.1357 & 42.3492 & 24.3542 & 35.2076 \\
        Gemini     & 31.2513 & 40.2169 & 26.3028 & 36.4054 \\
        Llama Merged & 66.4780 & 70.2607 & 51.7437 & \textbf{70.7144} \\
        Qwen Merged & \textbf{66.8993} & \textbf{70.9748} & \textbf{52.9495} & 61.4781 \\
        \bottomrule
    \end{tabular}
    }
\label{tab:eval_qwen_general}
\end{table}

The versions of the API models are: DeepSeek-V3-0324, moonshot-v1-8k, doubao-1-5-pro-32k-250115, gpt-4o-2024-08-06, gemini-2.5-flash. It is clear that the API tests of these general knowledge LLMs are all inferior to fine-tuned Qwen Merged model in the four evaluation indicators, which further demonstrate the effectiveness of the fine-tuning experiments.

\subsection{KG Construction Experiments}

\begin{figure}[H]
    \centering
    \includegraphics[width=0.9\linewidth]{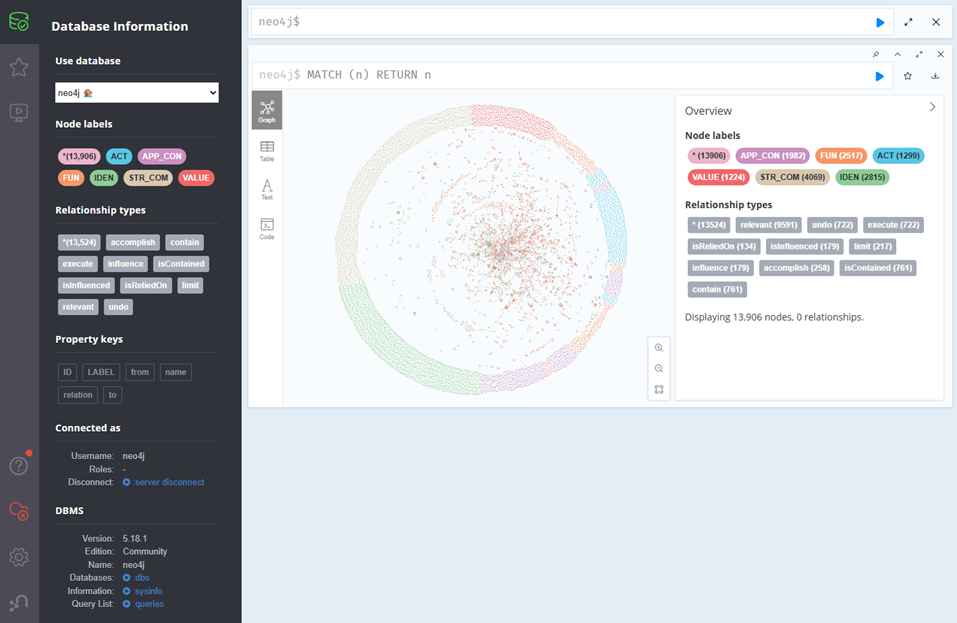} 
    \caption{The KG constructed.}
    \label{fig:kg_constructed}
\end{figure}

In the Neo4j platform, entities and relations were imported to display the KG constructed. There are 6 entity types and 10 relation types, with a total of 13,906 entity names and 13,524 relations. The results are shown in Fig. \ref{fig:kg_constructed}.

\subsection{QA System Experiments}

The aggregated results are shown in Table~\ref{tab:deepseek_eval}, the ``w/ KG-RAG'' results outperform the corresponding ``w/ Text-RAG'' and ``w/o KG-RAG'' results on all groups, which demonstrates the advantages of our KG-RAG pipeline. Especially in the experiment of the Qwen Merged group, the average score increases by 2.26\%, which confirmed that our KG2QA framework combines fine-tuning and KG-RAG pipeline, making stepwise improvements to the base model, and the answers given are more superior and reliable. It is clear that the improvement effect of our KG-RAG pipeline in the Base group is better than that in the Merged group of the same model. This may indicate that for models that have not been fine-tuned with domain-specific knowledge, the impact brought by using domain-specific KG is more obvious.

\begin{table}[!htbp]
\centering
\caption{Comparative evaluation results.}
\label{tab:deepseek_eval}
\resizebox{\linewidth}{!}{
\begin{tabular}{lcccccc}
\toprule
\textbf{Methods} & \textbf{Similarity} & \textbf{Fluency} & \textbf{Coherence} & \textbf{Relevance} & \textbf{Factual Acc.} & \textbf{Overall Avg.} \\
\midrule
\multicolumn{7}{c}{\bf Llama } \\
\midrule
Base w/o KG-RAG & 0.3321 & 0.8049 & 0.7513 & 0.7248 & 0.6177 & 0.6462 \\
Base w/ Text-RAG & 0.3684 & 0.8152 & 0.7631 & 0.7498 & 0.6459 & 0.6685 \\
Base w/ KG-RAG & 0.3854 & 0.8233 & 0.7706 & 0.7581 & 0.6593 & 0.6793 \\
\textbf{Base Improvement} & \textbf{+5.33\%} & \textbf{+1.84\%} & \textbf{+1.93\%} & \textbf{+3.33\%} & \textbf{+4.16\%} & \textbf{+3.31\%} \\
\addlinespace[0.5em]
\hdashline
\addlinespace[0.5em]
Merged w/o KG-RAG & 0.5190 & 0.9190 & 0.8620 & 0.8583 & 0.7684 & 0.7853 \\
Merged w/ Text-RAG & 0.5565 & 0.9201 & 0.8631 & 0.8715 & 0.7879 & 0.7998 \\
Merged w/ KG-RAG & 0.5681 & 0.9205 & 0.8642 & 0.8780 & 0.7953 & 0.8052 \\
\textbf{Merged Improvement} & \textbf{+4.91\%} & \textbf{+0.15\%} & \textbf{+0.22\%} & \textbf{+1.97\%} & \textbf{+2.69\%} & \textbf{+1.99\%} \\
\midrule
\multicolumn{7}{c}{\bf Qwen } \\
\midrule
Base w/o KG-RAG & 0.3416 & 0.8152 & 0.7639 & 0.7305 & 0.6284 & 0.6559 \\
Base w/ Text-RAG & 0.3805 & 0.8250 & 0.7750 & 0.7580 & 0.6550 & 0.6787 \\
Base w/ KG-RAG & 0.3927 & 0.8318 & 0.7811 & 0.7672 & 0.6658 & 0.6877 \\
\textbf{Base Improvement} & \textbf{+5.11\%} & \textbf{+1.66\%} & \textbf{+1.72\%} & \textbf{+3.67\%} & \textbf{+3.74\%} & \textbf{+3.18\%} \\
\addlinespace[0.5em]
\hdashline
\addlinespace[0.5em]
Merged w/o KG-RAG & 0.5278 & 0.9217 & 0.8673 & 0.8641 & 0.7728 & 0.7908 \\
Merged w/ Text-RAG & 0.5712 & 0.9225 & 0.8680 & 0.8794 & 0.7951 & 0.8072 \\
Merged w/ KG-RAG & 0.5836 & 0.9229 & 0.8695 & 0.8864 & 0.8045 & 0.8134 \\
\textbf{Merged Improvement} & \textbf{+5.58\%} & \textbf{+0.12\%} & \textbf{+0.22\%} & \textbf{+2.23\%} & \textbf{+3.17\%} & \textbf{+2.26\%} \\
\bottomrule
\end{tabular}
}
\end{table}

\begin{figure}[H]
    \centering
    \includegraphics[width=0.97\linewidth]{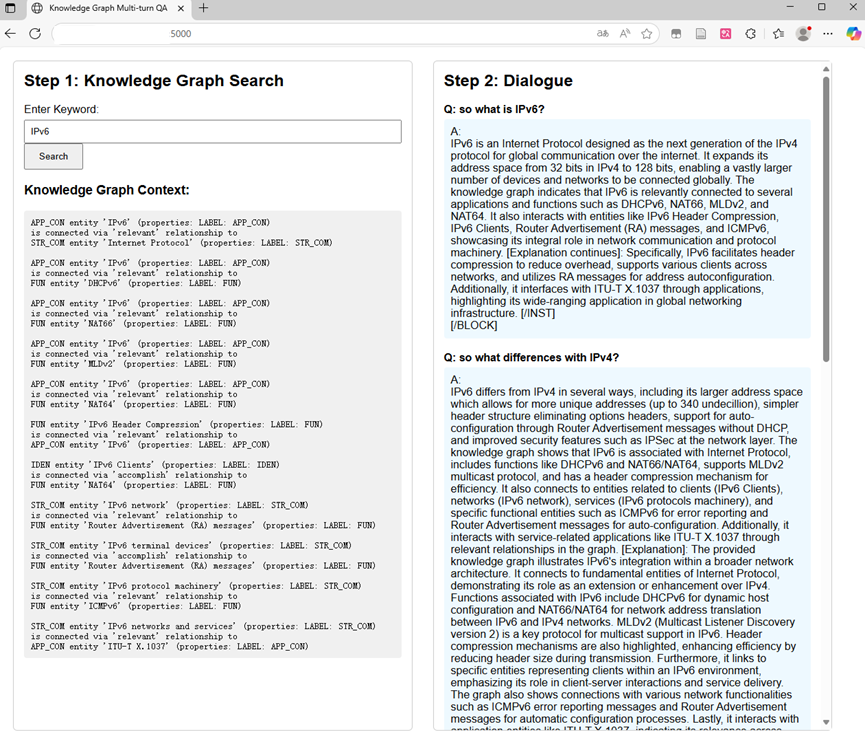} 
    \caption{The multi-round QA example of the Web platform.}
    \label{fig:qa_webpage}
\end{figure}

We use Flask to build the Web platform of the QA system and design two UI display parts on the left and right. Effectively display the two parts of the question and the answer in the left and right text boxes. We also utilize FastAPI to encapsulate the QA system as an API interface, providing Swagger interactive documentation and excellent performance support. We present an example of multi-round QA of ``IPv6''-related questions on the Web platform, as shown in Fig. \ref{fig:qa_webpage}. The system first retrieves and displays KG context for the keyword ``IPv6'', then uses this context to answer subsequent user questions about ``IPv6'' and its differences from ``IPv4''.

\section{Conclusion}
\label{sec:Con}

We propose the KG2QA framework and dataset for intelligent consulting services in the field of communication standards based on LLMs and KG, which improves the processing efficiency of consultation in the field of communication standards, reduces the burden of information search for communication engineers in their daily work, and has certain universality. This method can be extended and applied to the standardization consultation services of other specialties.

\bibliographystyle{IEEEbib}
\bibliography{icassp}

\vfill\pagebreak

© 20XX IEEE. Personal use of this material is permitted. Permission from IEEE must be obtained for all other uses, in any current or future media, including reprinting/republishing this material for advertising or promotional purposes, creating new collective works, for resale or redistribution to servers or lists, or reuse of any copyrighted component of this work in other works.

\end{document}